%% file: main.tex
\documentclass[11pt]{article}

\usepackage[final]{acl}

\usepackage{times}
\usepackage{latexsym}

\usepackage[T1]{fontenc}

\usepackage[utf8]{inputenc}

\usepackage{microtype}

\usepackage{inconsolata}

\usepackage{graphicx}
\usepackage{amsmath}
\usepackage{float}
\usepackage{url}

\usepackage{graphicx}
\usepackage{multirow}
\usepackage{booktabs}
\usepackage{url}
\usepackage{dblfloatfix}
\usepackage{array}
\usepackage{float}
\usepackage{xcolor}
\usepackage{pifont} 
\usepackage{amsmath}
\usepackage{amsfonts}

\newcommand{\cmark}{\textcolor{black}{\ding{51}}}

\usepackage{listings}
\usepackage{soul}
\usepackage{tabularx}
\usepackage{colortbl} 
\usepackage{paralist}
\usepackage{booktabs, array, multicol}
\usepackage{longtable}
\title{Accuracy and Satisfaction in Multi-Turn LLM Dialogues for NFR Assessment}

\author{Ali Pourghasemi Fatideh \\
  University of Maine \\
  \texttt{ali.pourghasemi@maine.edu} \\\And
  Wilder Baldwin \\
  University of Maine \\
  \texttt{wilder.baldwin@maine.edu} \\\And
  Maria Dhakal \\
  University of Notre Dame \\
  \texttt{mdhakal@nd.edu} \\\AND
  Collin McMillan \\
  University of Notre Dame \\
  \texttt{cmc@nd.edu} \\\And
  Sepideh Ghanavati \\
  University of Maine \\
  \texttt{sepideh.ghanavati@maine.edu} \\}

\begin{document}
\maketitle
\begin{abstract}
LLM-based dialogue assistants have become mainstream tools for software developers, yet current evaluation benchmarks focus exclusively on functional correctness. This leaves a critical gap in assessing the quality and accuracy of these conversations when handling Non-Functional Requirements (NFRs), which are inherently vague, context-dependent, and involve many parts of a program. Evaluating how well these systems support collaborative reasoning about NFRs requires methods that go beyond single-turn accuracy to capture both the correctness of the system's outputs and the quality of the multi-turn interaction.
In this paper, we investigate the accuracy and quality of multi-turn conversations between developers and an LLM-based agent in the domain of Health Insurance Portability and Accountability Act (HIPAA) regulatory compliance. 
We hired 49 programmers to interact with GitHub Copilot to assess 148 HIPAA-derived NFRs against the iTrust codebase, a system designed to comply with HIPAA regulations, across three dimensions: requirement satisfaction level, reasoning, and code localization.
We find that developers tend to agree with LLM assessments, but accuracy against expert ground truth is low. We model user satisfaction and find that longer system responses and more information-providing turns negatively affect user satisfaction, whereas proactive interactions positively affect it.
Our findings provide insights for designing LLM-based dialogue systems that support NFR assessment.
\end{abstract}

\input{Sections/introduction}
\input{Sections/approach}
\input{Sections/related_works_short}
\input{Sections/methodology}

\input{Sections/results}
\input{Sections/discussion}
\input{Sections/conclusions}
\input{Sections/limitations}
\input{Sections/ethics}
\input{Sections/acknowledgement}
\bibliography{custom}

\appendix
\input{Sections/appendix}

\end{document}

%% file: Sections/introduction.tex
\section{Introduction}
Interactive dialogue systems have demonstrated strong capabilities across various programming tasks \citep{fan2023large, ozkaya2023application, zhang2023survey}, including code generation, documentation, unit testing, fault and code localization, and bug fixing.
These capabilities have long been an aspiration in software engineering research, as programmers are most productive through collaborative problem-solving discussions with peers, yet such peer availability is often limited \citep{finch2020emora, gonzalez2004constant, latoza2006maintaining, perlow2002s, roehm2012professional, sillito2008asking}.
Recent advances in Large Language Models (LLMs) have enabled these practical interactive dialogue systems for developers.

As these AI systems engage with developers through multiple interaction turns, the success of the human-AI partnership hinges less on the raw correctness of any single generated code block and more on the quality of the collaborative problem-solving process \citep{kumar2025sharp, liu2024large}. This shift from evaluating individual outputs to assessing the collaborative dialogue marks an evolution in how we understand and measure AI assistance in software engineering \citep{kang2023large, meng2024empirical, nikolov2025google, rondon2025evaluating, vaithilingam2022expectation}. Yet despite this shift, current evaluation methods and benchmarks \citep{jimenez2023swe, jain2024livecodebench,aider2024polyglot} still focus primarily on single-turn functional correctness \citep{lin2025robunfr} and do not capture the accuracy or quality of multi-turn collaborative reasoning for Non-Functional Requirements (NFRs) \citep{arora2024advancing, gustavsson2024success}.
NFRs, unlike functional requirements (FRs), tend to be vaguely defined concerns that span many parts of a program \citep{glinz2005rethinking, mairiza2010investigation}, require domain knowledge, such as compliance with regulatory mandates, and may be difficult to test or have complex dependencies. Thus, they require evaluation techniques that could capture all the nuances regarding how well they are satisfied (as a property of the whole system) and their reasoning for such satisfactions. 

To this end, we investigate the accuracy and quality of multi-turn conversations in LLM-based Agents, with a focus on the Health Insurance Portability and  Accountability Act (HIPAA) \citep{hhs_hipaa} as a representative domain of regulatory NFRs. We construct a dataset of HIPAA-derived NFRs grounded in an Electronic Health Record (EHR) system and develop expert-annotated ground truth to capture the requirement satisfaction level, reasoning quality, and code location. We hire 49 programmers to interact with GitHub Copilot through multi-turn conversations and rate their agreement with the agent's assessments of NFRs. We then measure the accuracy of the LLM's assessments and identify which dialogue characteristics influence user satisfaction. Through this study, we address the following research questions:

\begin{compactitem}
    \item \textbf{RQ1:} How accurately do LLM-based dialogue assistants assess NFRs, and how do users perceive the response quality?
    \item \textbf{RQ2:} What dialogue characteristics significantly correlate with user satisfaction?
\end{compactitem}

Our findings show that participants agreed with the system's responses 91–94\% of the time across requirement satisfaction level, reasoning, and code location, yet accuracy against expert ground truth was low, with F1 scores of 0.381 for requirement satisfaction level, 0.520 (BERTScore) for reasoning, and F1 score of 0.203 for code location. 
These results suggest that LLM-based assistants produce responses that are perceived as high-quality but are inaccurate.
By modeling user satisfaction as a function of dialogue characteristics, we find that user satisfaction is correlated with three metrics. Verbose responses and a higher volume of information-providing turns are negatively associated with user satisfaction, whereas the number of proactive interaction turns is positively associated.
Our results provide insights for designing LLM-based dialogue systems that improve the developer experience in NFR assessment. Our data is open-source and available via: 
{\scriptsize\url{https://github.com/sh3rLock3d/DialogueNFR}}

%% file: Sections/approach.tex
\begin{figure*}[t]
  \includegraphics[width=0.96\linewidth]{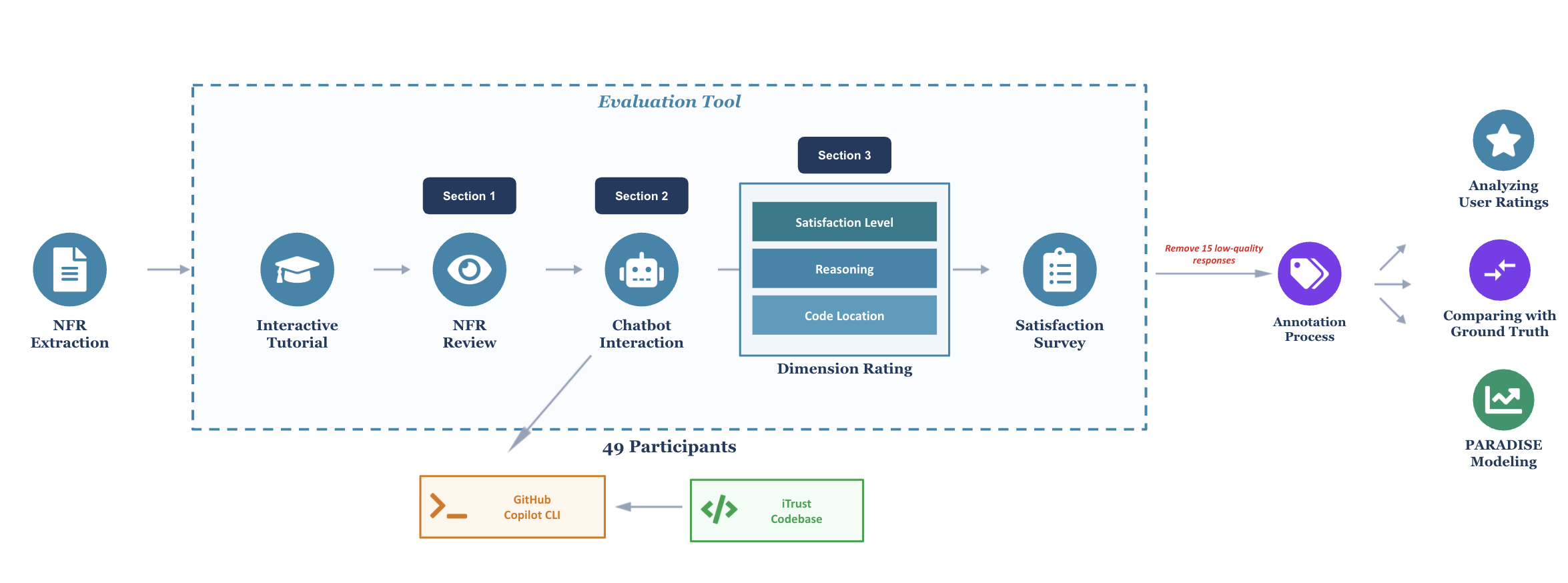}
  \caption {Methodology overview.}
  \label{fig:methodology_overview}
\end{figure*}

%% file: Sections/related_works_short.tex
\section{Related Work and Background}
This section discusses how our approach relates to prior work. Table~\ref{tab:related} summarizes prior work on evaluating LLMs.


\begin{table}[t]
\centering
\small
\caption{Summary of related work in evaluating LLMs.}
\label{tab:related}
\begin{tabular}{|l|c|c|l|}
\hline
\textbf{Work} & \textbf{MT} & \textbf{SM} & \textbf{Evaluation Target} \\
\hline
\citet{chen2021evaluating}   & -- & -- & Code generation \\
\citet{hendrycks2021measuring}   & -- & -- & Code generation \\
\citet{yin2018learning}   & -- & -- & Code generation \\
\citet{jimenez2023swe}   & -- & -- & Bug fixing \\
\citet{jain2024livecodebench}   & -- & -- & Code generation \\
\citet{aider2024polyglot}   & -- & -- & Code editing \\
\citet{wu2025humanevalcomm}   & -- & -- & Clarification ability \\
\citet{laban2025llms}   & \cmark & -- & MT degradation \\
\citet{vaithilingam2022expectation}   & \cmark & -- & User expectations \\
\citet{kumar2025sharp}   & \cmark & -- & Agent collaboration \\
\citet{liu2024refining}   & \cmark & -- & Code quality \\
\citet{sarkar2022like}   & \cmark & -- & User experience \\
\citet{liang2024large}   & \cmark & -- & Usability \\
\hline
\textit{This work}  & \cmark & \cmark & NFR assessment \\
\hline
\multicolumn{4}{l}{\scriptsize MT=Multi-turn, SM=Satisfaction Modeling} \\
\end{tabular}
\end{table}

\subsection{Evaluation of LLM-based Code Assistants}
The evaluation of LLM-based code assistants has been shaped by benchmarks targeting functional correctness, such as HumanEval \citep{chen2021evaluating}, APPS \citep{hendrycks2021measuring}, CoNaLa \citep{yin2018learning} and \citep{rodriguez2023benchmarking}, SWE-bench \citep{jimenez2023swe}, LiveCodeBench \citep{jain2024livecodebench}, and Aider Polyglot \citep{aider2024polyglot}. 
More recently, HumanEvalComm \citep{wu2025humanevalcomm} evaluated whether models ask clarifying questions when functional specifications are ambiguous. However, these benchmarks focus on whether models produce correct final outputs, rather than whether they can reason accurately over multiple turns.

\subsection{Dialogue System Evaluation}
The most widely adopted framework for task-oriented dialogue evaluation is PARADISE (PARAdigm for DIalogue System Evaluation) \citep{walker1997paradise}, which predicts overall user satisfaction as a function of task success and dialogue costs such as elapsed time and number of turns. PARADISE has been applied across diverse domains \citep{walker2002darpa,forbes2006modelling,bickmore2006health}, but to the best of our knowledge, no prior work has applied the PARADISE framework or similar user satisfaction-based dialogue evaluation methods to assess LLM-based dialogue systems.
With the rise of LLMs, automatic metrics like BERTScore \citep{zhang2019bertscore} or LLM-as-judge \citep{zhuge2024agent} have been proposed for utterance-level quality assessment. Yet these metrics evaluate static output quality and do not model how dialogue characteristics influence user satisfaction. \citet{laban2025llms} further show that LLM performance degrades in multi-turn conversations, which reinforces the need for evaluation methods that go beyond single-turn quality.
In this work, we apply the PARADISE framework to an LLM-based dialogue system and assess the final response obtained through multi-turn collaboration.

\subsection{Developer--AI Interaction Studies}
A growing body of work examines how developers interact with AI coding tools. \citet{vaithilingam2022expectation} found gaps between developer expectations and actual experience with LLM-generated code, while \citet{kumar2025sharp} studied how developers wield agentic AI in real software engineering tasks. Studies have identified issues with overconfidence \citep{liu2024refining}, poor intent specification \citep{sarkar2022like}, and usability limitations \citep{liang2024large} that functional benchmarks cannot capture. However, these works fall short of identifying which LLM behaviors or dialogue characteristics influence developer satisfaction.

\subsection{iTrust}
HIPAA is a U.S. federal regulation that establishes standards for protecting the privacy and security of individually identifiable health information, known as electronic protected health information (ePHI). 
We use HIPAA as a representative domain of regulatory NFRs.
To provide a realistic codebase for NFR assessment, we use an EHR system called iTrust \citep{meneely2012appendix}. The goal of iTrust is to provide students with a project that is real-world relevant and sufficiently deep and complex to simulate industrial systems.
iTrust is well-suited for this study, as it serves as a testbed for understanding security and privacy requirements, with a focus on HIPAA rules, making it an appropriate context for evaluating HIPAA-derived NFRs.

%% file: Sections/methodology.tex
\section{Methodology}
Figure~\ref{fig:methodology_overview} illustrates our methodology. We first extract 148 HIPAA-derived NFRs and construct ground truth assessments against the iTrust codebase. We then recruit 49 programmers to use our evaluation tool, where they review the NFRs, interact with a GitHub Copilot agent to assess them, and rate the agent's responses.
They then complete a user satisfaction survey.  We annotate the dialogue turns based on dialogue characteristics.

\subsection{Ground Truth Creation}
\label{subsec:ground-truth}
To construct the NFR corpus, a privacy and software engineering expert reviewed the HIPAA regulations and extracted 650 statements. Next, to ensure that the NFRs are traceable in iTrust, we applied two inclusion criteria: (1) the statement must pertain to only software construction defined in SWEBOK \citep{bourque1999guide}, and (2) it must fall within iTrust's scope.
These criteria reduced 650 statements to 180 NFRs. Then, these NFRs were reviewed by another software engineering expert to ensure they satisfied the criteria and met the characteristics of a good NFR as defined by Sommerville \citep{sommerville2011software}. The other reviewer proposed some edits to the NFR texts and identified 32 NFRs that remain outside the scope of iTrust. These steps reduced 650 statements to the final list of 148 NFRs, covering topics such as technical safeguards and use of ePHI (e.g., \textit{The system shall ensure the integrity of ePHI that it creates.}).

As mentioned earlier, our goal is to evaluate whether LLMs can provide an analyses that match human understanding. Specifically: what is the compliance status (requirement satisfaction level), why does that status hold (reasoning), and where in the code is the requirement addressed (localization).
We then manually create a ground truth to compare LLM responses and assess the LLM's accuracy and quality (RQ1). This ground truth includes an answer for each NFR across three dimensions.
These three dimensions are (i) requirement satisfaction Level, which has four values as Satisfied, Weakly Satisfied, Weakly Denied, or Denied \citep{amyot2010evaluating}; (ii) reasoning behind the satisfaction level; and (iii) code locations that address the NFR, including file name and line number(s).
We choose these dimensions because, for finding code location related to an NFR, binary links (i.e., related vs. non-related) do not capture how code implements, or relates to requirements \citep{alor2025evaluating}. 

To construct the ground truth, the first expert assessed each NFR with respect to iTrust, documenting the satisfaction level, reasoning behind their assessment, and relevant code locations. A second expert then reviewed each annotation to verify its correctness or provide feedback on any issues identified. The first expert either accepted the feedback or, in cases of disagreement, engaged in discussion with the second expert to reconcile differing assessments. Discussions continued until both experts reached an agreement, and the agreed-upon annotations were recorded as the ground truth.

\subsection{Survey Design}
\label{sec:survey-design}
We design a survey to evaluate LLM performance in assessing NFRs through multi-turn dialogue. In this survey, developers are asked to review a subset of NFRs, query an LLM-based chatbot to assess NFRs, rate the chatbot’s responses, and finally complete a user satisfaction survey (adopted from the PARADISE framework \citep{walker1997paradise}).

Before starting the surveys, participants complete an interactive tutorial that walks them through the codebase structure and asks them to evaluate three NFRs. 
Upon successful completion of the tutorial, they proceed with the main survey, which has three stages.
In the first stage, developers review 10 assigned NFRs to familiarize themselves with the requirements before interacting with the chatbot. For each NFR, they need to acknowledge that they reviewed it. We also include two attention-check questions to ensure developers read the NFRs carefully.
In the second stage, developers query the chatbot to assess the NFRs across the three dimensions.
We ask developers to query all NFRs in a single initial prompt, and then follow up based on the agent's initial response to elicit more targeted subsequent prompts.
In the third stage, developers rate each NFR's responses on the three dimensions on a Likert scale of 1–4, where 1 is "Strongly Disagree" and 4 is "Strongly Agree". We exclude a neutral option to nudge developers to provide a rating \citep{jain2026automated,su2024distilled}; however, we provide a text box where developers can explain if they are unable to make a selection. Here, we also include two  additional attention-check questions to ensure participants do not respond randomly.
After completing the main survey, they are asked to complete a user satisfaction survey and a demographic questionnaire. The satisfaction survey contains questions adopted from the PARADISE framework, 
which we use to model the system's performance. 
We ensured that each NFR is assigned to at least two developers.

\subsection{Survey Tool}
\begin{figure}[ht]
  \includegraphics[width=0.96\linewidth]{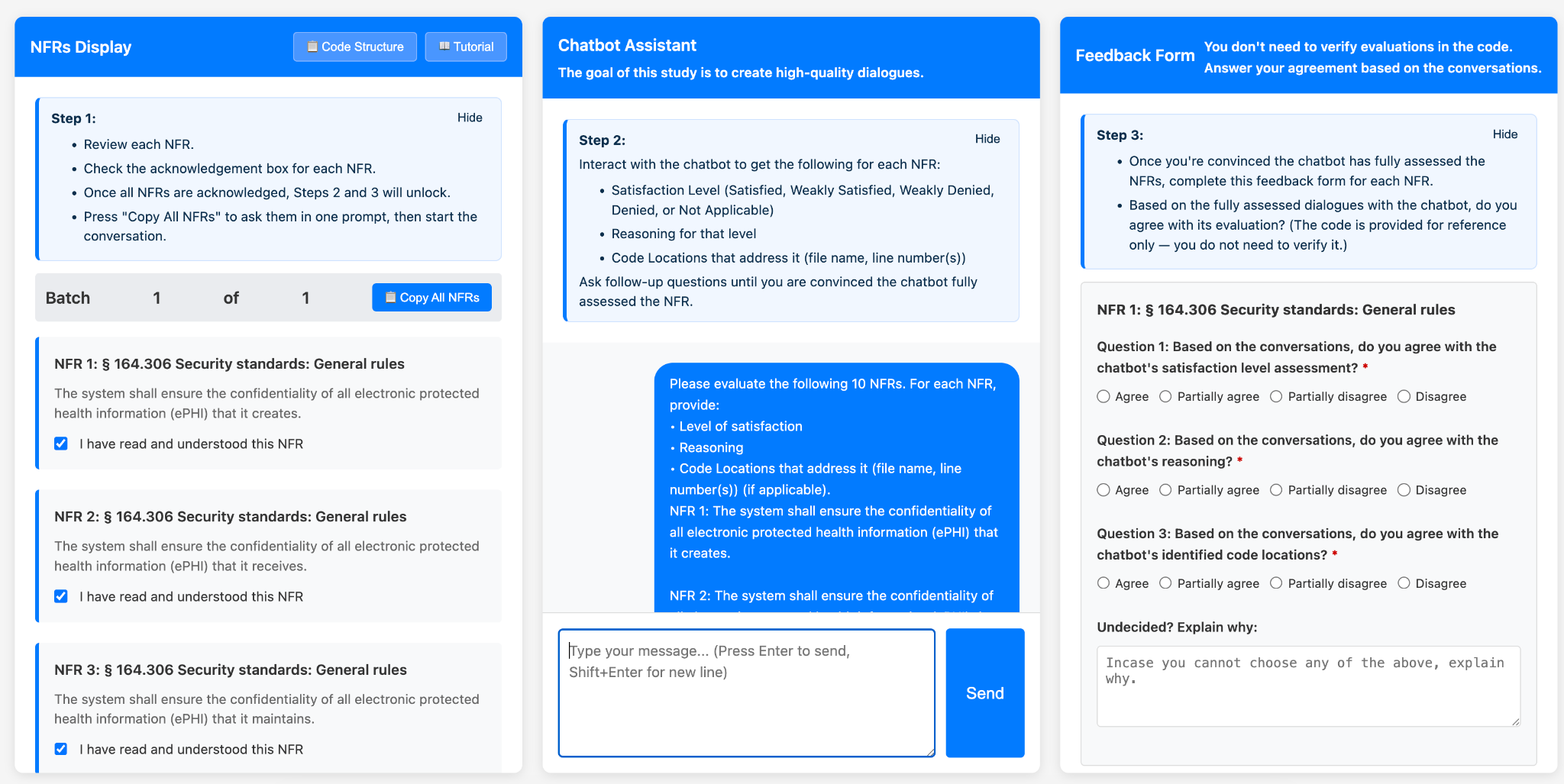}
  \caption {User interface of the tool: review NFRs (left), chatbot interaction (middle), and response rating  (right).}
  \label{fig:website_overview}
\end{figure}

We developed a tool to support NFR review, chatbot interaction, and response rating. 
Figure~\ref{fig:website_overview} shows our tool, which consists of three panels (one for each stage described in Section \ref{sec:survey-design}).

The chatbot is powered by GitHub Copilot, and we use the GitHub Copilot CLI \citep{github_copilot_cli_2026} to relay the agent's responses to developers. We use gpt-5.1-codex-max \citep{openai_gpt51codexmax} as the underlying model. 

\subsection{Participants}
We first conducted a pilot study with 8 participants: 4 computer science students from our university and 4 recruited through Prolific \citep{prolific}, to validate the study design. The pilot confirmed that the task could be completed in a reasonable time (6 participants completed it within an hour) and that the instructions were clear (everyone who interacted with the chatbot successfully completed the study). 
For the main study, we recruited 41 participants through Prolific. We required participants to hold a degree in Computer Science or Mathematics, work in a software engineering-related role, and have an approval rating of 95–100\% on the platform.
Pilot and main study participants were compensated \$15 and \$20, respectively. This study was approved by our Institutional Review Board (IRB) (see Section \ref{sec:ethics} for more details).
After screening for quality, we removed 15 responses in which participants failed the attention-check questions, user turns were unrelated to the given NFRs, or the dialogues lacked a satisfaction level, reasoning, or code location, resulting in 34 participants.

\subsection{Annotation Process}
To analyze dialogue characteristics that correlate with user satisfaction (RQ2), we annotate each turn in the collected dialogues using labels drawn from three sources: the PARADISE framework, which provides dialogue cost metrics, MT-Eval \citep{kwan2024mt}, and MT-Bench-101 \citep{bai2024mt}, which show the interaction patterns in the turns. Appendix~\ref{sec:annotation-def} provides a definition of the annotation labels used in this study.

Two experts independently reviewed a subset of 50 turns and developed heuristics for annotation based on these taxonomies. In the first round of annotation, the average inter-rater reliability calculated using Cohen's Kappa \citep{cohen1960coefficient} between the two annotators was 0.32, indicating fair agreement. The annotators then resolved disagreements and refined the annotation heuristics. Using the updated heuristics, they re-annotated the same set and achieved a Cohen’s kappa of 0.83, indicating almost perfect agreement. After resolving the disagreements, a next round of 54 additional turns yielded an average Cohen's kappa of 0.81, indicating almost perfect agreement again. Given this high level of agreement, the two annotators independently annotated the remaining turns.

\begin{table}[t]
  \centering
  \small
  \begin{tabular}{lccc}
    \hline
    \textbf{Label} & \textbf{R1} & \textbf{R1-redo} & \textbf{R2} \\
    \hline
    Recollections            & 0.000 & 0.640 & 0.772 \\
    Expansion                & 0.129 & 0.602 & 0.666 \\
    Refinement               & 0.176 & 0.729 & 0.615 \\
    Follow-up                & 0.289 & 0.796 & 0.884 \\
    user initiatives         & 0.668 & 0.898 & 1.000 \\
    unsuitable-request       & --    & --    & --    \\
    inappropriate responses  & --    & --    & --    \\
    Error                    & --    & --    & --    \\
    help messages            & 0.000 & --    & --    \\
    given-data               & 0.408 & 1.000 & -- \\
    check move               & 0.369 & 0.638 & 0.791 \\
    relevant data            & --    & --    & --    \\
    abandoned requests       & 0.000 & 1.000 & -- \\
    Anaphora Resolution      & 0.000 & 1.000 & 1.000 \\
    Separate Input           & --    & --    & --    \\
    Content Confusion        & --    & --    \\
    Content Rephrasing       & 0.370 & 0.912 & 0.813 \\
    Format Rephrasing        & 0.000 & --    & --    \\
    Self-correction          & 0.558 & 0.811 & -- \\
    Self-affirmation         & 0.728 & 0.878 & 0.769 \\
    Instruction Clarification & 0.479 & 0.790 & -- \\
    Proactive Interaction    & 1.000 & 1.000 & -- \\
    \hline
    \textbf{Size}            & 50 & 50    & 54    \\
    \textbf{Average $\kappa$} & 0.323 & 0.836 & 0.812 \\
    \hline
  \end{tabular}
  \caption{Inter-rater reliability for dialogue turn labels across annotation rounds. Labels with – indicate insufficient variation in annotations to compute $\kappa$.}
  \label{tab:kappa}
\end{table}


%% file: Sections/results.tex
\section{Results}
In this section, we present the results of our study. We outline the basic structure and descriptive statistics of the collected dialogues, and examine the accuracy and perceived quality of the LLM-based assistant's NFR assessments. We then apply the PARADISE framework to identify which dialogue characteristics significantly influence user satisfaction.
Table~\ref{tab:dialogue_stats} summarizes the key statistics of the collected dialogues.


\begin{table}
  \centering
  \small
  \begin{tabular}{lr}
    \hline
    \textbf{Statistic} & \textbf{Value} \\
    \hline
    Total \# Dialogues          & 34     \\
    Total \# Turns              & 406    \\
    Avg. \# Turns per Dialogue  & 11.94  \\
    Avg. \# Words in Prompt     & 60.90  \\
    Max. \# Words in Prompt     & 496    \\
    Avg. \# Words in Response   & 129.46 \\
    Max. \# Words in Response   & 877    \\
    Avg. \# Words per Turn      & 190.36 \\
    Max. \# Words per Turn      & 1206   \\
    \hline
  \end{tabular}
  \caption{Key statistics of the collected dialogues.}
  \label{tab:dialogue_stats}
\end{table}

\subsection{RQ1: Response Accuracy and Quality}
We evaluate the system's NFR assessments from two perspectives. First, we measure accuracy by comparing the system's outputs against the ground truth in Section~\ref{subsec:ground-truth}. Second, we measure response quality through participant agreement ratings done in stage three of the survey.

\subsubsection{Ground Truth Accuracy}
\label{subsec:accuracy}
To measure the accuracy of the system's outputs, we manually extracted the system's final responses on the evaluation dimensions from each dialogue and compared them against the ground truth.

We compute accuracy differently for each dimension, due to the nature of each output. Satisfaction level is a categorical label, so we compute precision, recall, and F1 score. Reasoning is free-form text, so we measure semantic similarity between the system's reasoning and the ground truth using BERTScore \citep{zhang2019bertscore}, report mean precision, recall, and F1 across all responses, and additionally report cosine similarity and ROUGE \citep{lin2004rouge} as complementary measures of semantic alignment. Code location consists of sets of file names and line numbers; we define true positives as $|G \cap P|$, false positives as $|P \setminus G|$, and false negatives as $|G \setminus P|$, where $G$ is the ground truth set and $P$ is the system's predicted set, and compute the mean precision, recall, and F1 across NFRs.

We present these results in Table~\ref{tab:accuracy}.
The system achieved an F1 score of 0.381 for satisfaction level, only modestly above the 0.25 expected from random assignment across four categories. This suggests that while the system can produce responses with high perceived quality, it struggles to correctly determine the satisfaction status of NFRs.
For reasoning, BERTScore F1 is 0.520, cosine similarity is 0.774, and ROUGE-1/2/L F1 scores are 0.204, 0.037, and 0.156, respectively.
The relatively higher cosine similarity is expected, as both the system and ground truth address the same NFR and are in the same topical space. However, the lower BERTScore indicates that the generated reasoning diverges from the ground truth.
Code location yields a mean F1 of 0.203, indicating that the system also struggles to identify where in the codebase NFRs are addressed.

\begin{table}
\centering
\small
\begin{tabular}{llr}
\toprule
\textbf{Dimension} & \textbf{Metric} & \textbf{Score} \\
\midrule
Satisfaction Level & Precision & 0.438 \\
Satisfaction Level & Recall & 0.418  \\
Satisfaction Level & F1 & 0.381 \\
\midrule
\multirow{5}{*}{Reasoning} 
& Precision & 0.521 \\
& Recall &  0.529 \\
& F1 & 0.520 \\
 & Cosine Similarity & 0.774 \\
 & ROUGE-1 & 0.204 \\
 & ROUGE-2 & 0.037 \\
 & ROUGE-L & 0.156 \\
\midrule
Code Location & Precision & 0.151 \\ 
Code Location & Recall & 0.142 \\ 
Code Location & F1 & 0.203 \\ 
\bottomrule
\end{tabular}
  \caption{Accuracy of LLM against ground truth.}
   \label{tab:accuracy}
\end{table}


\subsubsection{Participant Perceived Response Quality}

\begin{figure}[t]
  \includegraphics[width=\columnwidth]{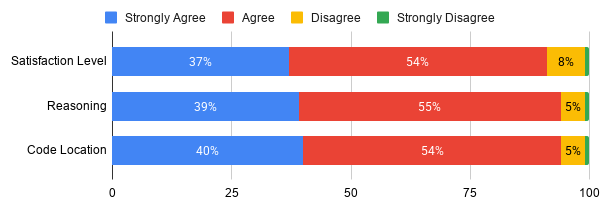}
  \caption{Distribution of participant agreement ratings across the three evaluation dimensions.}
  \label{fig:rating_distribution}
\end{figure}

\paragraph{Distribution of Ratings.} Figure~\ref{fig:rating_distribution} shows the distribution of participant ratings across the three dimensions. The majority of participants agreed or strongly agreed with the system's responses: 91\% for satisfaction level, 94\% for reasoning, and 94\% for code location. Disagreement rates were low, at 8\% for satisfaction level, 6\% for reasoning, and 5\% for code location. "Strongly disagree" ratings were rare, with only 4, 5, and 2 instances for satisfaction level, reasoning, and code location, respectively. Mean agreement was 3.27 for satisfaction level, 3.31 for reasoning, and 3.34 for code location, all above the ``agree'' threshold (Table~\ref{tab:agreement}). To assess whether mean agreement was significantly above the ``agree'' threshold of 3, we conducted one-sample $t$-tests with $H_0$: $\mu \leq 3$ versus $H_1$: $\mu > 3$. All three dimensions were statistically significant at $p < 0.001$. These results confirm that participants' rating of the system's responses is significantly above the ``agree'' threshold across all three dimensions, indicating that users generally perceive response quality as high. Table \ref{tab:ttest} in Appendix~\ref{sec:ttest} reports the full test statistics. These results stand in contrast to the low accuracy, suggesting that the system produces responses perceived as high-quality by developers but are inaccurate.

\paragraph{Rating Divergence.} 
To identify NFRs where participants diverged in their ratings, we computed the Average Deviation (AD) index \citep{burke1999average} for each NFR across participants and flagged those with $AD > 0.67$, the recommended cutoff for 4-point Likert scales indicating a lack of consensus \citep{burke2002estimating}. 
We adopt AD because, unlike other agreement metrics, it does not require specification of a null distribution, thereby avoiding assumptions about the nature of random responding \citep{o2017overview}. This is particularly important for our data, where ratings are skewed toward "agree" and "strongly agree," making metrics that assume balanced distributions less appropriate.
Of the 148 NFRs, 19 exhibited high divergence on satisfaction level, 13 on reasoning, and 15 on code location (Table~\ref{tab:agreement}). 
Of these, 6 NFRs had high divergence on all three dimensions and 6 on two dimensions. Among single-dimension divergences, code location accounted for 9 cases, satisfaction level for 7, and reasoning for only 1. 
These results suggest that system quality is less consistent across satisfaction levels for the same NFR than other dimensions.


\begin{table}
  \centering
  \small
  \begin{tabular}{lcc}
    \hline
    \textbf{Dimension} & \textbf{Mean Agr.} & \textbf{High AD} \\
    \hline
    Satisfaction Level & 3.27 & 19 \\
    Reasoning          & 3.31 & 13 \\
    Code Location      & 3.34 & 15 \\
    \hline
  \end{tabular}
  \caption{Mean participant agreement and number of NFRs where participant rating diverged ($AD > 0.67$).}
  \label{tab:agreement}
\end{table}




\subsection{RQ2: Dialogue Characteristics and User Satisfaction}
To investigate which dialogue characteristics influence user satisfaction, we applied the PARADISE framework, following the methodological considerations outlined by \citet{hajdinjak2006paradise} and using predictors derived from annotations.

\begin{figure}[t]
  \includegraphics[width=\columnwidth]{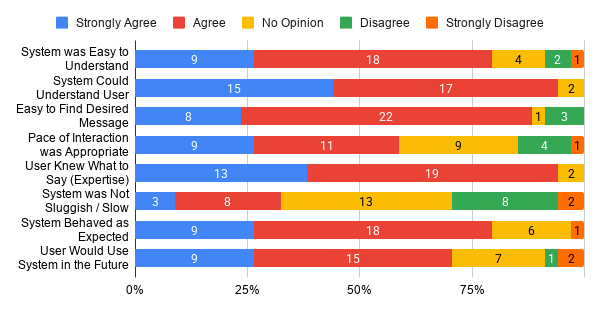}
  \caption{Participants' PARADISE ratings for the LLM.}
  \label{fig:user_satisfaction}
\end{figure}

\paragraph{User Satisfaction.} After completing the main evaluation, participants filled out a satisfaction survey adopted from the PARADISE framework.
Figure~\ref{fig:user_satisfaction} summarizes the distribution of user satisfaction ratings. The user satisfaction score (US), computed as the sum of individual Likert responses, ranged from 20 to 37, with a mean of 31.26 out of 40, indicating that users were generally satisfied with the system.
The features related to the system's ability to understand the user and the user's ability to know what to say at each point were the most highly rated, averaging 4.3. The features related to the system's speed received the lowest ratings, averaging around 3.
These ratings suggest that the system created a fluent and intuitive conversational experience, which may have contributed to the high agreement rates observed in RQ1.

\begin{table}[ht]
  \centering
  \small
  \begin{tabular}{lr}
    \hline
    \textbf{Label} & \textbf{Count} \\
    \hline
    Given Data              & 370 \\
    Follow-up                    & 195 \\
    Recollections                & 156 \\
    Expansion                    & 152 \\
    Refinement                   &  53 \\
    User Initiatives        &  37 \\
    Proactive Interaction        &  36 \\
    Content Rephrasing           &  29 \\
    Check Move (CM)              &  27 \\
    Self-affirmation             &  23 \\
    Self-correction              &  16 \\
    Separate Input               &  11 \\
    Instruction Clarification    &  10 \\
    Anaphora Resolution          &   9 \\
    Unsuitable Request        &   3 \\
    Abandoned Requests       &   2 \\
    Help Messages             &   2 \\
    Format Rephrasing            &   2 \\
    Inappropriate Responses   &   1 \\
    Error                        &   1 \\
    Content Confusion            &   0 \\
    Relevant Data             &   0 \\
    \hline
  \end{tabular}
  \caption{Total counts of dialogue act labels.}
  \label{tab:label_counts}
\end{table}

\paragraph{All Predictors.}
We construct a set of predictors that may correlate with user satisfaction. These predictors include task success, task completion, dialogue costs \citep{hajdinjak2006paradise}, and patterns \citep{kwan2024mt,bai2024mt}.
Task success is computed as the sum of F1 scores across the three evaluation dimensions described in Section~\ref{subsec:accuracy}. 
Task completion is computed as the proportion of NFRs for which the system provided all three outputs, except when the NFR is assessed as Denied, in which case we require only requirement satisfaction level and reasoning, since code location is not applicable. 
Dialogue costs and patterns can be obtained by metadata (e.g., elapsed time) and annotated data. Table~\ref{tab:label_counts} summarizes the results of annotated data.

In Table~\ref{tab:predictors} in Appendix~\ref{sec:predictors}, we report the descriptive statistics for all predictor variables. We first removed predictors with zero variance (\textit{content confusion}, \textit{relevant data count} and \textit{ratio}), as these provide no discriminative information. We then identified highly correlated predictor pairs (i.e., the absolute values of the correlation coefficients $|r| > 0.7$) and, for each pair, retained the predictor with the lower p-value against US. This is because with highly correlated predictors, small errors or variations in the observed values can lead to large changes in the weights of the performance function \citet{hajdinjak2006paradise}. 
In Table~\ref{tab:removed_corr} in Appendix~\ref{sec:predictors}, we show the 9 removed predictors and their correlated counterparts, leaving 21 predictors for the regression. We use these 21 predictors as candidate dialogue characteristics that may correlate with user satisfaction.


\paragraph{Predictors of User Satisfaction.}
We analyzed which predictors significantly correlate with user satisfaction and can be used to estimate it. To this end, we developed a performance function for the LLM system that predicts user satisfaction based on the predictors. The process for creating this function can help us identify the predictors that are significantly correlated  \citet{hajdinjak2006paradise}.
Before creating the function, we removed outliers based on deviations from the mean. One dialogue was excluded because its user satisfaction value exceeded two standard deviations from the mean. Additionally, one dialogue was removed due to an extreme predictor value, with one predictor exceeding five standard deviations from the mean. After removing outliers, 32 dialogues remained in the dataset. We then z-score-normalized all variables to ensure that the resulting regression coefficients reflect each predictor's relative contribution to user satisfaction. 
We fit the PARADISE performance function using backward elimination \citep{seber2003linear} and iteratively removing the least significant predictor until all remaining predictors were significant. The full elimination sequence is reported in Table~\ref{tab:backward} in Appendix~\ref{sec:backward}. The elimination process removed 18 predictors. The final model retained three significant predictors, which result in the following performance function:
\vspace{-2mm}
\[
\begin{aligned}
\hat{N}(\text{US})
&= -0.583 \cdot N(\text{Mean Words per Response}) \\[-0ex]
&\phantom{=}\; -0.485 \cdot N(\text{Given Data}) \\[-0ex]
&\phantom{=}\; +0.422 \cdot N(\text{Proactive Interaction})
\end{aligned}
\]

In this function, $N$ indicates the normalized value of the metric. The results show that user satisfaction is significantly influenced by only these dialogue characteristics.
We report the coefficient details in Table~\ref{tab:paradise}. The negative coefficients for \textit{mean bot words} ($\beta = -0.583$) and \textit{given data count} ($\beta = -0.485$) suggest that verbose responses are associated with lower user satisfaction, even when the information itself may be relevant. In contrast, \textit{proactive interaction} ($\beta = 0.422$) as the only positive predictor, suggests that participants value the system taking the initiative in the dialogue and contribute positively to user satisfaction.


\begin{table}
  \centering
  \small
  \setlength{\tabcolsep}{4pt}
  \begin{tabular}{lrrrr}
    \hline
    \textbf{Predictor} & $\beta$ & \textbf{SE} & $t$ & $p$ \\
    \hline
    Mean Words per Response        & $-0.583$ & 0.174 & $-3.35$ & .002 \\
    Given data            & $-0.485$ & 0.162 & $-2.99$ & .006 \\
    Proactive Interaction &  $0.422$ & 0.163 &  $2.60$ & .015 \\
    \hline
    \multicolumn{4}{l}{\scriptsize $\beta$: standardized coefficient; SE: standard error; $t$: test statistic.} \\
  \end{tabular}
  \vspace{-1em}
  \caption{PARADISE performance function coefficients.}
  \label{tab:paradise}
\end{table}

%% file: Sections/discussion.tex
\section{Discussion}

\subsection{The Gap between Rating and Accuracy}
LLM-based code assistants have demonstrated strong performance on functional correctness \citep{chen2021evaluating, li2022competition, jain2024livecodebench}. However, our results reveal that this competence does not extend to NFRs. 
Participants agreed or strongly agreed with the system's outputs over 91\% of the time, yet F1 scores against expert ground truth were low.
As an example, an NFR requires the system to implement procedures for creating passwords. The ground truth provides its reasoning based on dedicated creation actions. However, the system reasons based on \texttt{ResetPasswordAction.java} as evidence and conflates password reset with password creation. 
Here, the system failed to discern semantically similar yet functionally distinct operations. Despite this mistake, the system produced a sufficiently convincing response that the participant strongly agreed with the incorrect assessment.
This example shows that while LLMs can generate convincing NFR assessments that satisfy developers, their actual accuracy remains low. 
Several factors, such as LLM's confidence in responses, NFRs' attributes (e,g,. ambiguity), or lack of expertise may have contributed to this gap.
Understanding precisely which factors drive the gap, and how to mitigate them, is a direction for future work.

\subsection{Dialogue Design Implications}
By applying the PARADISE framework, we derived a function that measures the performance of LLM agents for evaluating NFRs. This function provides actionable insights for developing effective LLM-based dialogue assistants.
Beyond measuring the performance, our data enables a further step toward fully automated evaluation. Creating a dialogue system still requires human participants to interact with the system, whereas user simulation can automate the evaluation process entirely. Our annotated dialogue corpus provides the foundation for training user simulators that can evaluate LLM-based agents without requiring human participants, which we plan to pursue in future work.



%% file: Sections/conclusions.tex
\section{Conclusions}
This study investigated multi-turn dialogues between developers and an LLM-based agent for evaluating NFRs. Our results reveal that participants agreed with the system's outputs over 91\% of the time, yet actual accuracy was low (F1 of 0.381 for satisfaction level, 0.520 for reasoning, and 0.203 for code localization). By applying the PARADISE framework, we found that a high number of proactive interactions positively influence user satisfaction, while verbose responses and a high number of information-providing responses detract from it. These findings provide guidance on how LLM-based dialogue agents should interact with developers in NFR assessment tasks.

%% file: Sections/limitations.tex
\section{Limitations}
This study has several limitations that should be considered when interpreting the results.

First, the NFR corpus was initially extracted by a single privacy and software engineering expert who reviewed the HIPAA regulations and derived 180 NFRs from 650 extracted statements. A second expert reviewed and validated these extracted NFRs, which raises the possibility that some relevant requirements were missed. HIPAA is a complex and extensive regulatory framework, and a single reviewer may overlook statements that a second independent extractor would have identified. In the future, we will ensure that two experts independently extract NFRs from the regulations and reconcile their results to ensure a more comprehensive coverage of the regulatory space.

Second, we evaluated a single LLM-based agent (GitHub Copilot with gpt-5.1-codex-max) against a single codebase (iTrust) in one regulatory domain (HIPAA). While iTrust is a testbed for privacy and security requirements, the generalizability of our findings to other codebases, regulatory frameworks, or LLM-based agents remains an open question.


Third, like most survey-based studies in software engineering, this work faces several validity threats. Variations in the source code and NFRs, participant selection, or the survey interface could affect outcomes. To address these concerns, we use the iTrust EHR system, a well-established codebase specifically designed to comply with HIPAA regulations, from which we draw our NFRs, and a simple web interface to avoid the complexity of tools like IDEs. 
Another problem is that our participant pool, recruited through Prolific with screening filters to target individuals with software development experience, may not fully represent all developers who engage with NFR assessment in practice or have knowledge in HIPAA compliance, and may be subject to selection bias.
The study is also subject to self-report bias and recall bias, as participants rated the system's responses based on their own understanding of HIPAA compliance, which may not reflect expert-level judgment, and may have misremembered details of the interaction when completing the satisfaction survey.


Fourth, our analysis was conducted on a relatively small dataset, i.e., 32 dialogues, which limits the statistical power of the analysis. Although we applied standard procedures, including predictor filtering, normalization, and backward elimination, the identified predictors and their coefficients may not be stable or generalizable beyond this dataset.

Finally, our user satisfaction function should not be used to increase or optimize user satisfaction in an LLM agent, because it may make inaccurate responses seem more convincing.



%% file: Sections/ethics.tex
\section{Ethics Consideration}
\label{sec:ethics}
This study was approved by the University of Maine's Institutional Review Board (IRB) and conducted in compliance with institutional ethical standards. Prior to participation, all individuals were provided informed consent through a form detailing the investigators, risks, benefits, compensation, and confidentiality. They were informed that participation was voluntary, and individuals were free to withdraw at any point. Contact information for the investigators and the IRB was provided in the consent form. No participants contacted either party during the study.

%% file: Sections/acknowledgement.tex
\section{Acknowledgement}
\label{sec:ack}
This research was supported by NSF Award \#2442683.

%% file: Sections/appendix.tex
\section{Participants Demographic}
\label{sec:participants-demographic}

\begin{figure*}
  \includegraphics[width=0.48\linewidth]{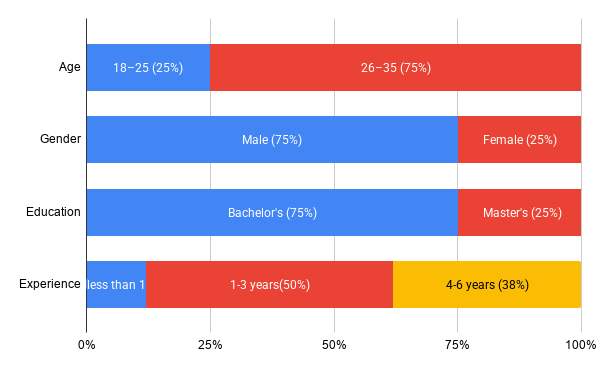} \hfill
  \includegraphics[width=0.48\linewidth]{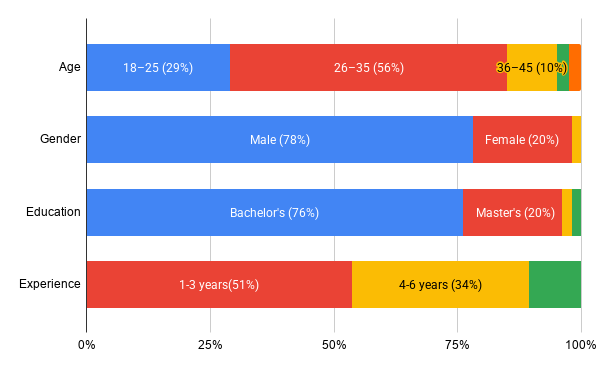}
  \caption{Demographic distribution of pilot study participants (n=8, left) and main study participants (n=41, right).}
  \label{fig:demographics}
\end{figure*}

We recruited a total of 49 participants for our study: 8 from a pilot study and 41 from the main study. Figure~\ref{fig:demographics} shows the demographic distribution of participants in each group.

\section{Annotation Definitions}
\label{sec:annotation-def}

Table~\ref{tab:annotation-schema} defines the annotation labels used in this study, along with their definitions and examples from our corpus.

\section{Statistical Significance Tests}
\label{sec:ttest}

In Table~\ref{tab:ttest}, we report the one-sample $t$-test results for participant agreement across the three evaluation dimensions.
\begin{table}[h]
\small
  \centering
  \begin{tabular}{lcccc}
    \hline
    \textbf{Dimension} & \textbf{Mean} & \textbf{SD} & $t$ & $p$ \\
    \hline
    Satisfaction Level & 3.27 & 0.66 & 7.49  & $<.001$ \\
    Reasoning          & 3.31 & 0.63 & 9.05  & $<.001$ \\
    Code Location      & 3.34 & 0.60 & 10.41 & $<.001$ \\
    \hline
  \end{tabular}
  \caption{One-sample $t$-test results for participant agreement ($H_0$: $\mu \leq 3$).}
  \label{tab:ttest}
\end{table}

\begin{table}[t]
  \centering
  \small
  \begin{tabular}{llr}
    \hline
    \textbf{Removed} & \textbf{Retained} & $r$ \\
    \hline
    Errors                    & Inappropriate Resp. & 1.000 \\
    Inappropriate Resp. Ratio & Inappropriate Resp. & 1.000 \\
    Abandoned Requests        & Abandoned Req. Ratio & 0.999 \\
    Help Message Ratio        & Help Messages        & 0.996 \\
    Number of Turns           & Given data           & 0.981 \\
    Expansion                 & Given data           & 0.867 \\
    Follow-up                 & Given data           & 0.860 \\
    Recollections             & Given data           & 0.853 \\
    Unsuitable Req. Ratio     & Unsuitable Requests  & 0.806 \\
    \hline
  \end{tabular}
  \caption{Predictors removed due to high correlation ($|r| > 0.7$). For each pair, the predictor with the higher $p$-value against US was removed.}
  \label{tab:removed_corr}
\end{table}

\begin{table}[t]
  \centering
  \small
  \begin{tabular}{clcc}
    \hline
    \textbf{Step} & \textbf{Removed Predictor} & $p$ & $R^2$ \\
    \hline
     1 & Mean Words Request        & 0.853 & 0.728 \\
     2 & Given data Ratio          & 0.802 & 0.727 \\
     3 & Mean Response Time        & 0.865 & 0.726 \\
     4 & Task Completion           & 0.803 & 0.725 \\
     5 & Instruction Clarification & 0.725 & 0.724 \\
     6 & Self-correction           & 0.681 & 0.721 \\
     7 & Check Move Ratio          & 0.652 & 0.718 \\
     8 & Check Moves               & 0.688 & 0.714 \\
    9 & Unsuitable Requests       & 0.534 & 0.711 \\
    10 & Inappropriate Responses   & 0.484 & 0.705 \\
    11 & Anaphora Resolution       & 0.446 & 0.697 \\
    12 & User Initiatives          & 0.138 & 0.688 \\
    13 & Elapsed Time              & 0.074 & 0.652 \\
    14 & Refinement                & 0.111 & 0.597 \\
    15 & Help Messages             & 0.150 & 0.549 \\
    16 & Self-affirmation          & 0.157 & 0.507 \\
    17 & Abandoned Request Ratio   & 0.166 & 0.465 \\
    18 & Task Success              & 0.134 & 0.423 \\
    \hline
  \end{tabular}
  \caption{Backward elimination steps. At each step, the predictor with the highest $p$-value was removed. The process terminated when all remaining predictors were significant ($p < 0.05$).}
  \label{tab:backward}
\end{table}

\section{Predictors}
\label{sec:predictors}
In Table~\ref{tab:predictors}, we report the descriptive statistics for user satisfaction and all predictor variables used in the PARADISE regression model.

\begin{table*}[t]
  \centering
  \small
  \begin{tabular}{lrrrr}
    \hline
    \textbf{Variable} & \textbf{Mean} & \textbf{SD} & \textbf{Min} & \textbf{Max} \\
    \hline
    User Satisfaction (US)                    & 31.26 &    3.65 &   20.00 &    37.00 \\
    \hline
    \multicolumn{5}{l}{\textit{Dialogue Cost Metrics - meta-data \citep{hajdinjak2006paradise}}} \\
    Number of Turns                           & 11.94 &    7.95 &    2.00 &    27.00 \\
    Mean Response Time (s)                    & 45.80 &   33.05 &   12.59 &   157.77 \\
    Elapsed Time (s)                            & 3496.42 & 9230.81 & 312.92 & 55369.17 \\
    Mean Words per Request                       & 60.90 &   42.13 &    9.41 &   229.50 \\
    Mean Words per Response                            & 129.46 &  105.49 &   50.91 &   548.00 \\
    \hline
    \multicolumn{5}{l}{\textit{Dialogue Cost Metrics - Counts \citep{hajdinjak2006paradise}}} \\
    User Initiatives                          &  1.09 &    0.38 &    1.00 &     3.00 \\
    Unsuitable Requests                       &  0.09 &    0.29 &    0.00 &     1.00 \\
    Inappropriate Responses                   &  0.03 &    0.17 &    0.00 &     1.00 \\
    Errors                                    &  0.03 &    0.17 &    0.00 &     1.00 \\
    Help Messages                             &  0.06 &    0.24 &    0.00 &     1.00 \\
    Given data                                & 10.88 &    7.47 &    2.00 &    26.00 \\
    Check Moves                               &  0.79 &    1.09 &    0.00 &     4.00 \\
    Relevant Data                             &  0.00 &    0.00 &    0.00 &     0.00 \\
    Abandoned Requests                        &  0.06 &    0.24 &    0.00 &     1.00 \\
    \hline
    \multicolumn{5}{l}{\textit{Dialogue Cost Metrics - Ratios \citep{hajdinjak2006paradise}}} \\
    Unsuitable Request Ratio                  &  0.009 &   0.036 &   0.000 &    0.200 \\
    Inappropriate Response Ratio              &  0.004 &   0.021 &   0.000 &    0.125 \\
    Help Message Ratio                        &  0.003 &   0.014 &   0.000 &    0.063 \\
    Given data Ratio                          &  0.907 &   0.121 &   0.583 &    1.000 \\
    Check Move Ratio                          &  0.064 &   0.105 &   0.000 &    0.500 \\
    Relevant Data Ratio                       &  0.000 &   0.000 &   0.000 &    0.000 \\
    Abandoned Request Ratio                   &  0.002 &   0.009 &   0.000 &    0.040 \\
    \hline
    \multicolumn{5}{l}{\textit{Interaction Pattern Labels \citep{kwan2024mt,bai2024mt}}} \\
    Recollections                             &  4.59 &    3.64 &    0.00 &    11.00 \\
    Expansion                                 &  4.47 &    4.01 &    0.00 &    12.00 \\
    Refinement                                &  1.56 &    2.27 &    0.00 &    10.00 \\
    Follow-up                                 &  5.74 &    5.48 &    0.00 &    18.00 \\
    Anaphora Resolution                       &  0.26 &    0.57 &    0.00 &     2.00 \\
    Content Confusion                         &  0.00 &    0.00 &    0.00 &     0.00 \\
    Self-correction                           &  0.47 &    1.11 &    0.00 &     5.00 \\
    Self-affirmation                          &  0.68 &    1.39 &    0.00 &     5.00 \\
    Instruction Clarification                 &  0.29 &    0.52 &    0.00 &     2.00 \\
    Proactive Interaction                     &  1.06 &    2.17 &    0.00 &    12.00 \\
    \hline
    \multicolumn{5}{l}{\textit{Task Metrics}} \\
    Task Success                              &  0.95 &    0.26 &    0.55 &     1.80 \\
    Task Completion                           &  0.90 &    0.20 &    0.10 &     1.00 \\
    \hline
  \end{tabular}
  \caption{Descriptive statistics for user satisfaction and all predictor.}
  \label{tab:predictors}
\end{table*}




\section{High Correlated Predictors}
\label{sec:high_corr}
Table~\ref{tab:removed_corr} lists the predictors removed due to high correlation ($|r| > 0.7$)

\section{Backward Elimination Steps}
\label{sec:backward}
In Table~\ref{tab:backward}, we report the full backward elimination sequence.

\onecolumn
 
\begin{longtable}{|p{2.5cm}|p{4.5cm}|p{7cm}|}
\caption{Annotation schema with definitions and examples from our study.}
\label{tab:annotation-schema} \\
 
\hline
\textbf{Metric} & \textbf{Definition} & \textbf{Notes / Example} \\
\hline
\endfirsthead
 
\multicolumn{3}{l}{\textit{Table~\ref{tab:annotation-schema} (continued)}} \\
\hline
\textbf{Metric} & \textbf{Definition} & \textbf{Notes / Example} \\
\hline
\endhead
 
\hline
\multicolumn{3}{r}{\textit{Continued on next page}} \\
\endfoot
 
\hline
\endlastfoot
Recollections & User returns to a topic or detail from earlier in the dialogue (not the immediately preceding turn), requiring the system to retrieve information from its prior context. & Applies when the user explicitly navigates back to a previously discussed topic. E.g., \textit{``Ok, back to NFR1: What exactly constitutes a `discrepancy' in login attempts---e.g., multiple failures from one IP, geolocation mismatches, or failed 2FA?''} \\
\hline
Expansion & User explores a new facet or sub-aspect of the current topic without changing to a different topic entirely. & Often occurs when the user has already asked about one dimension (e.g., satisfaction level) and then asks about another (e.g., code location) for the same NFR. E.g., \textit{``What about access of authorized software programs?''} (a new facet of NFR~12 after discussing human-role access). May co-occur with Follow-up. \\
\hline
Refinement & User modifies, narrows, or corrects their previous instructions after seeing the system's output. & The user reviews the output and asks for a changed version. May trigger Content Rephrasing, Format Rephrasing, or Self-correction on the system side. E.g., \textit{``Shouldn't the office visit details be covered within `patient records'? Dig deeper into office visit details to clarify this.''} \\
\hline
Follow-up & User asks a question that builds directly on the system's most recent response, referencing specific details or opinions from it. & The key criterion is \textit{most recent response}. If the user refers back to an earlier turn, annotate as Recollections instead. E.g., \textit{``Where in code can this be found?''} (referencing the role-enforcement explanation just given) \\
\hline
User Initiative & User starts a new information-seeking request or introduces a new topic. & Includes the first turn of a session as well as mid-conversation topic shifts (e.g., when the user moves from one NFR to the next). E.g., \textit{``NFR~5: The system shall support implementing policies...''} (introducing a new NFR for evaluation) \\
\hline
Unsuitable Request & User's request is out of scope or context of the system's domain. & No instances in our corpus. \\
\hline
Inappropriate Response & System produces an unexpected or erroneous response, including pardon moves. & No instances in our corpus. \\
\hline
Error & System errors, including malformed or incoherent natural-language output. & No instances in our corpus. \\
\hline
Help Message & System proactively provides guidance, recommendations, or actionable steps beyond what was directly requested. & Distinct from Given Data: GD answers the question asked; HM goes further by offering unsolicited help. E.g., user asks \textit{``Why is NFR~38 denied? Indicate where this check needs to occur.''} and the system not only explains the gap but recommends where authorization checks should be added. \\
\hline
Given Data & System provides the information the user directly requested. & Present in the majority of turns. E.g., system returns satisfaction level, reasoning, and code locations for each NFR in the batch. \\
\hline
Check Move & User or system requests verification or confirmation of something stated in a prior turn. Extended from the original PARADISE definition (system-only) to include user-initiated checks. & E.g., \textit{``Can you validate this again to make sure the roles are correct?''} or \textit{``So nothing was accomplished at this NFR, right? Not a single thing.''} or user challenges: \textit{``Are you sure that is true?''} \\
\hline
Relevant Data & System directs the user to select from relevant, available data. & No instances in our corpus. \\
\hline
Abandoned Request & User drops a request without receiving a satisfying answer and moves on. & E.g., \textit{``Let's start again from 0: Please evaluate the following 10 NFRs...''} (abandoning the prior single-NFR approach) \\
\hline
Anaphora Resolution & The user's query contains a pronoun or demonstrative whose referent must be resolved from prior turns. & E.g., \textit{``How is this varified?''} (``this'' refers to ePHI integrity from the prior turn). Also: \textit{``I have the same question for NFR~39.''} (``the same question'' refers to the NFR~38 query pattern) \\
\hline
Separate Input & Task requirements and the actual input data are provided across different turns rather than in a single message. & E.g., Turn~1: user establishes evaluation criteria (satisfaction level, reasoning, code locations). Turn~7: \textit{``With the experience evaluating NFR~11, analyze these NFRs as well. [NFR~12--20]''} \\
\hline
Content Confusion & User's query conflates superficially similar but semantically distinct concepts, requiring the system to disambiguate. & E.g., \textit{``There are encryption terms for the user authentication, check again.''} (user conflates password hashing with ePHI encryption; system correctly distinguishes SHA-256 hashing from data-at-rest encryption) \\
\hline
Content Rephrasing & System rephrases the content of its previous response to provide more depth or detail, triggered by user feedback. & Describes the system's response; often co-occurs with Refinement on the user side. E.g., \textit{``I want Reasoning to be a little more detailed, thank you.''} System regenerates with expanded justifications. \\
\hline
Format Rephrasing & System restructures the format or presentation of its previous response without changing the substance, triggered by user feedback. & Describes the system's response; often co-occurs with Refinement on the user side. E.g., \textit{``Ok but you didn't provide line numbers for those files.''} and \textit{``Nope, line number. I need exact code not just file names.''} \\
\hline
Self-correction & System revises its previous answer based on the user's feedback or challenge. & E.g., user asks \textit{``So is that NFR satisfaction level correct for NFR~12 with the part of the authorized software programs?''} System downgrades NFR~12 from Satisfied to Weakly Satisfied. \\
\hline
Self-affirmation & System maintains its previous position when challenged by the user. & E.g., user asks \textit{``Is this not just about getting the information during an emergency for the ER roles?''} System maintains Weakly Satisfied, explaining the gap between ER-role access and broader emergency override. \\
\hline
Instruction Clarification & System asks questions because the user's request is too ambiguous to proceed without more information. & E.g., user says \textit{``Ok but that's not implemented, I don't see the point to that.''} System responds: \textit{``Can you tell me what behavior you're seeing? Are you testing through the normal /login.jsp flow or elsewhere?''} \\
\hline
Proactive Interaction & System voluntarily asks probing questions to deepen the conversation, even though it could answer without them. & Distinct from Instruction Clarification: the system does not \textit{need} the information to proceed but asks to encourage exploration. E.g., after listing missing PHI coverage areas, the system asks: \textit{``Which PHI areas matter most to you so I can dig deeper?''} \\

\end{longtable}
 
\twocolumn